\documentclass[sigconf]{acmart}

\makeatletter
\def\@ACM@checkaffil{
    \if@ACM@instpresent\else
    \ClassWarningNoLine{\@classname}{No institution present for an affiliation}%
    \fi
    \if@ACM@citypresent\else
    \ClassWarningNoLine{\@classname}{No city present for an affiliation}%
    \fi
    \if@ACM@countrypresent\else
        \ClassWarningNoLine{\@classname}{No country present for an affiliation}%
    \fi
}
\makeatother

\usepackage{amssymb}
\usepackage{amsmath,amsfonts}
\usepackage{algorithmic}
\usepackage{array}
\usepackage[caption=false,font=normalsize,labelfont=sf,textfont=sf]{subfig}
\usepackage{textcomp}
\usepackage{stfloats}
\usepackage{url}
\usepackage{verbatim}
\usepackage{graphicx}
\usepackage{algorithm}
\usepackage{bbm}
\usepackage{bm}
\usepackage{ulem}
\usepackage{arydshln}
\usepackage{enumitem}

\usepackage{algorithmic}
\usepackage{amsfonts,amssymb} 
\usepackage{amsmath}
\usepackage{multicol}
\usepackage{multirow} 
\usepackage{stfloats}
\usepackage{color}
\usepackage{balance}
\usepackage[mathscr]{euscript}
\usepackage{hyperref}
\hypersetup{hidelinks,
	colorlinks=true,
	allcolors=blue,
	pdfstartview=Fit,
	breaklinks=true}
 \newcommand{\tabincell}[2]{\begin{tabular}{@{}#1@{}}#2\end{tabular}} 
\AtBeginDocument{%
  }

\setcopyright{acmcopyright}
\copyrightyear{2023}
\acmYear{2023}
\setcopyright{acmlicensed}\acmConference[MM '23]{Proceedings of the 31st ACM International Conference on Multimedia}{October 29-November 3, 2023}{Ottawa, ON, Canada}
\acmBooktitle{Proceedings of the 31st ACM International Conference on Multimedia (MM '23), October 29-November 3, 2023, Ottawa, ON, Canada}
\acmPrice{15.00}
\acmDOI{10.1145/3581783.3612532}
\acmISBN{979-8-4007-0108-5/23/10}

\begin{document}

\title{Dynamic Compositional Graph Convolutional Network for Efficient Composite Human Motion Prediction}
\author{Wanying Zhang}
\orcid{0009-0001-6387-6769}
\affiliation{%
  \institution{Sun Yat-Sen University}}

\author{Shen Zhao}
\orcid{0000-0002-4698-2658}
\authornote{Corresponding authors. \\
Shen Zhao: z-s-06@163.com;  \\
Mengyuan Liu: nkliuyifang@gmail.com}
\affiliation{%
  \institution{Sun Yat-Sen University}}

\author{Fanyang Meng}
\orcid{0000-0001-5725-2178}
\affiliation{%
  \institution{Peng Cheng Laboratory}
}

\author{Songtao Wu}
\orcid{0000-0003-1578-8980}
\affiliation{%
 \institution{Sony R\&D Center China, Beijing Lab}}

\author{Mengyuan Liu}
\orcid{0000-0002-6332-8316}
\authornotemark[1]
\affiliation{%
  \institution{Key Laboratory of Machine Perception, Shenzhen Graduate School, Peking University}}

\renewcommand{\shortauthors}{Wanying Zhang, Shen Zhao, Fanyang Meng, Songtao Wu, \& Mengyuan Liu}

%

\begin{abstract}
With potential applications in fields including intelligent surveillance and human-robot interaction, the human motion prediction task has become a hot research topic and also has achieved high success, especially using the recent Graph Convolutional Network (GCN). Current human motion prediction task usually focuses on predicting human motions for atomic actions. Observing that atomic actions can happen at the same time and thus formulating the composite actions, we propose the composite human motion prediction task. To handle this task, we first present a Composite Action Generation (CAG) module to generate synthetic composite actions for training, thus avoiding the laborious work of collecting composite action samples. Moreover, we alleviate the effect of composite actions on demand for a more complicated model by presenting a Dynamic Compositional Graph Convolutional Network (DC-GCN). Extensive experiments on the Human3.6M dataset and our newly collected CHAMP dataset consistently verify the efficiency of our DC-GCN method, which achieves state-of-the-art motion prediction accuracies and meanwhile needs few extra computational costs than traditional GCN-based human motion methods. Our code and dataset are publicly available at \href{https://github.com/Oliviazwy/DCGCN}{https://github.com/WanyingZhang/DCGCN} 
\end{abstract}




\maketitle


\section{Introduction}
Human motion prediction refers to predicting future poses based on observations of historical postures. This task has a wide range of applications in areas such as autonomous driving, intelligent surveillance, and human-robot interaction \citep{zhu2020topology,zhu2021recurrent,liu2017enhanced, wang2021spatio,liu2018recognizing}. 
Current methods \citep{mao2019learning,li2020predicting,corona2020context,mao2020history,aksan2021spatio,dang2021msr,ma2022progressively} usually focus on human motion prediction for atomic actions. These atomic actions are commonly collected to describe daily life actions. Taking the most popular Human3.6M dataset \citep{ionescu2013human3} for human motion prediction as an example, 15 types of atomic actions such as ``walking" and ``eating" are used in the training and testing stages.

We observe that some atomic actions can happen at the same time. For example, one can perform ``waving" meanwhile ``walking". This type of ``waving while walking" is called composite action which contains multiple atomic actions.
This common phenomenon motivates us to investigate the task of composite human motion prediction.
To ensure the generalization ability, a composite human motion prediction algorithm is expected to handle both composite actions and also traditional atomic actions. 

\begin{figure}[!ht]
\centering 
\includegraphics[width=0.99\columnwidth]{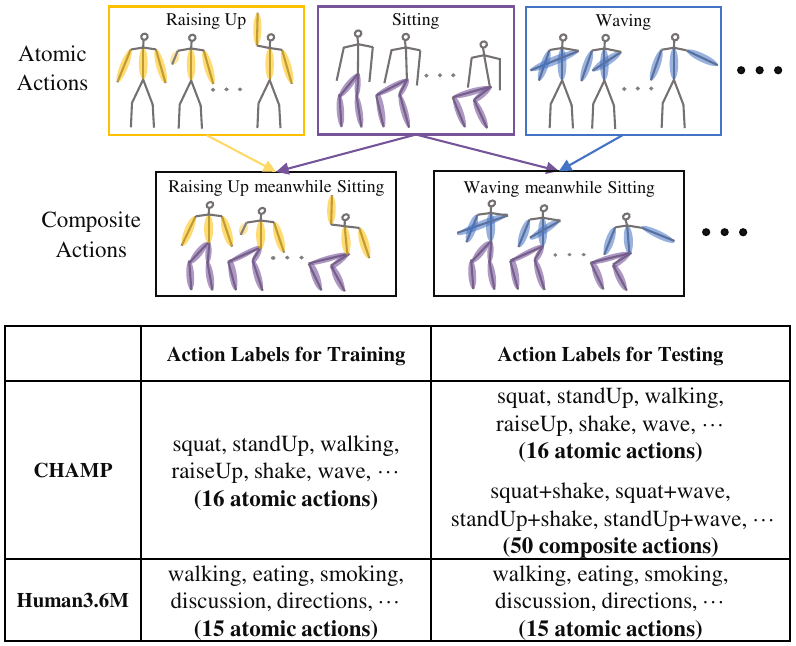}
\vspace{-0.5em}
\caption{\uline{In contrast to the current human motion prediction task that uses the same atomic action category for training and testing, the composite human motion prediction task involves not only atomic actions but also composite actions in the testing phase}. The image illustrates that atomic actions occurring in different human body parts are combined to form composite actions. The table compares the action labels assigned to the training and test datasets of the mainstream dataset (e.g., Human3.6M \citep{ionescu2013human3}) for the human motion prediction task and our proposed CHAMP dataset for the composite human motion prediction task.}
\vspace{-0.5em}
\label{tasks}
\end{figure}

To handle this task, a natural question first arises: \textbf{How to efficiently collect composite action samples for training?} Compared with atomic actions, direct collecting similar scales of composite actions is time-consuming and laborious. The reason is that the potential number of composite actions is extremely larger than that of atomic actions, as different combinations of atomic actions can formulate different composite actions. \uline{We avoid the problem of direct collecting composite actions with a Composite Action Generation (CAG) module, that uses atomic actions to generate synthetic composite actions}. Specifically, our proposed CAG module uses solely atomic actions as training data to synthesize composite actions. This module, based on VAE \citep{kingma2013auto}, tries to reconstruct the input atomic actions during the training stage to enable the model to discern the characteristics of each atomic action. During the generation stage, multiple atomic actions are simultaneously fed into the model, and the masking mechanism is employed to merge and transform these atomic actions into corresponding composite actions. Since current human motion prediction datasets barely contain composite actions, we collect a new Composite HumAn Motion Prediction dataset, called CHAMP dataset, to benchmark the composite human motion prediction task. As shown in Figure \ref{tasks}, our proposed CHAMP dataset contains 16 types of atomic actions for training, as well as 16 types of atomic actions and 50 types of composite actions for testing.

Besides, another natural question is: \textbf{How to design an efficient composite human motion prediction network?}
Compared with traditional human motion prediction network that only needs to handle atomic actions, the composite human motion prediction network has to retain the ability to predict human motions for both atomic actions and more complicated composite actions.
\uline{We alleviate the effect of composite actions on model design by using a Dynamic Compositional Graph Convolutional Network (DC-GCN), which integrates early exit mechanisms with predictors and uses policy networks that allow the model to adaptively learn which layer to output for each sample}. This mechanism ensures both prediction accuracy for complex samples and reduces redundant computation for simple samples during the inference stage, thus improving computational efficiency. Additionally, after analyzing the characteristics of the data, we identify significant differences in the complexity of various actions, with atomic actions generally being simpler than composite actions. We also observe that the complexity of the same action varies across different body parts of the human body. For instance, the action of ``hand-waving'' primarily occurs in the upper body. Based on these data features, we design a Compositional GCN as a predictor. To avoid redundancy in training atomic actions, we design multiple branches in the predictor to model different body parts. However, for composite actions involving the entire body skeleton, dividing them into parts for modeling would reduce body consistency. Therefore, an additional branch is implemented to model the entire human skeletal structure.
Our main contributions are three-fold.

\begin{itemize}[leftmargin=*]
\item Compared with the current human motion prediction task, we present a more practical composite human motion prediction task, which uses atomic actions as training data and aims to predict human motions for both atomic actions and composite actions that consists of multiple atomic actions. The current human motion prediction task can be treated as a special case of our proposed composite human motion prediction task.

\item We collect a new Composite HumAn Motion Prediction dataset, called the CHAMP dataset, which contains 16 types of atomic actions for training meanwhile contains 16 types of atomic actions and 50 types of composite actions for testing. Our proposed CHAMP dataset serves as a benchmark for the composite human motion prediction task.

\item We present a Composite Action Generation (CAG) module, that uses atomic actions to generate synthetic composite actions.
Taking advantage of synthetic samples generated by our CAG module, we further present a Dynamic Compositional Graph Convolutional Network (DC-GCN) for composite human motion prediction.
Extensive experiments on Human3.6M dataset and our CHAMP dataset verify that our DC-GCN can achieve state-of-the-art prediction accuracies meanwhile needs few extra computational costs than traditional GCN-based human motion prediction networks.

\end{itemize}

\section{RELATED WORK}

\quad \textbf{Human Motion Prediction. }
Recently, GCN-based methods have had a great performance in human perception \citep{cui2021towards,dang2021msr, li2020dynamic, mao2019learning,vaswani2017attention,mao2020history,mao2021multi}, regarding the structure of the human skeleton as a graph. \citet{mao2019learning} use a feed-forward graph convolutional network with learnable adjacent matrices in human motion prediction. The attitude is regarded as a fully connected graph and GCN is used to model the relationship between any pair of joints. The joint trajectories are represented by Discrete Cosine Transform (DCT) coefficients in the temporal domain. Recently, \citet{guo2023back} propose a light-weight-network based on simple Multi-Layer Perceptrons, which has an outstanding performance in human motion prediction with only 0.14 million parameters. Although the latest research uses Multi-Layer Perceptrons to predict human body movements, it has limitations in encoding spatial information. Therefore, we choose GCN as our base model for its ability to effectively model joint relationships and spatial dependencies.

\textbf{Motion Synthesis.} The insufficiency of training data is a universal challenge in human motion prediction tasks including the composite human motion prediction task, and data augmentation is one approach to address this challenge. The term “data augmentation” comes from \citet{tanner1987calculation}, associating the augmented data with the observed data by a many-to-one mapping. Classical data augmentation methods are usually based on transformation, such as cropping, flipping, rotating, adding random noise, scaling, random warping, etc. 
Gaussian noise is a simple but effective approach for data augmentation, where \citet{fragkiadaki2015recurrent} increase the variety of input data by destroying the input motion using zero mean Gaussian noise and \citet{lopes2019improving} add Gaussian noise to prevent the model from overfitting \citep{iwana2021empirical}. 
Color adjustment, blurring, sharpening, white balance, and other distortions have also been used in Image data augmentation \citep{krizhevsky2017imagenet,zoph2018learning}. Several conventional data augmentation techniques are applicable for skeleton-based motion synthesis, including fragmenting different parts of the skeleton from distinct movements and concatenating them. However, the cut-and-piece method is too crude and might synthesize physically-implausible motions. \citet{maeda2022motionaug} present a data augmentation scheme for human motion prediction consisting of Variational AutoEncoder (VAE) and Inverse Kinematics (IK), with motion correction using physics simulation to rectify unrealistic artifacts in the synthesized motions. 
Because of the favorable training stability of VAE, we constructed a VAE-based model as the Action Composite Module.

\textbf{Dynamic Networks.} The majority of popular deep learning models rely on a static inference approach, where the same set of models and parameters are utilized for all inputs. However, this approach may limit their expressiveness, efficiency, and interpretability \citep{graves2016adaptive,huang2017multi}. Alternatively, dynamic networks can adaptively assign different structures and parameters based on different input samples. As different input samples may have different computational requirements, it is reasonable to dynamically adjust the width and depth of the network. The mechanism of ``early exit'' allows simple samples to be exported at the shallower exit of the model without executing deeper layers \citep{huang2017multi,teerapittayanon2016branchynet,bolukbasi2017adaptive}, which not only saves redundant computation for simple examples but also maintains its representation ability when identifying difficult samples. 
Besides, the exit policies enable dynamic networks to make data-dependent decisions during inference to make samples exit at appropriate layers. The confidence-based criteria \citep{huang2017multi} and policy networks \citep{mcgill2017deciding,jie2019anytime} are commonly seen as decision-making schemes. We integrate the early exit mechanism into DC-GCN and employ several policy modules to determine where to exit samples on different branches in an adaptive manner. 
These modifications allow for efficient inference.

\begin{figure*}[t]
\centering 
\includegraphics[width=0.99\textwidth]{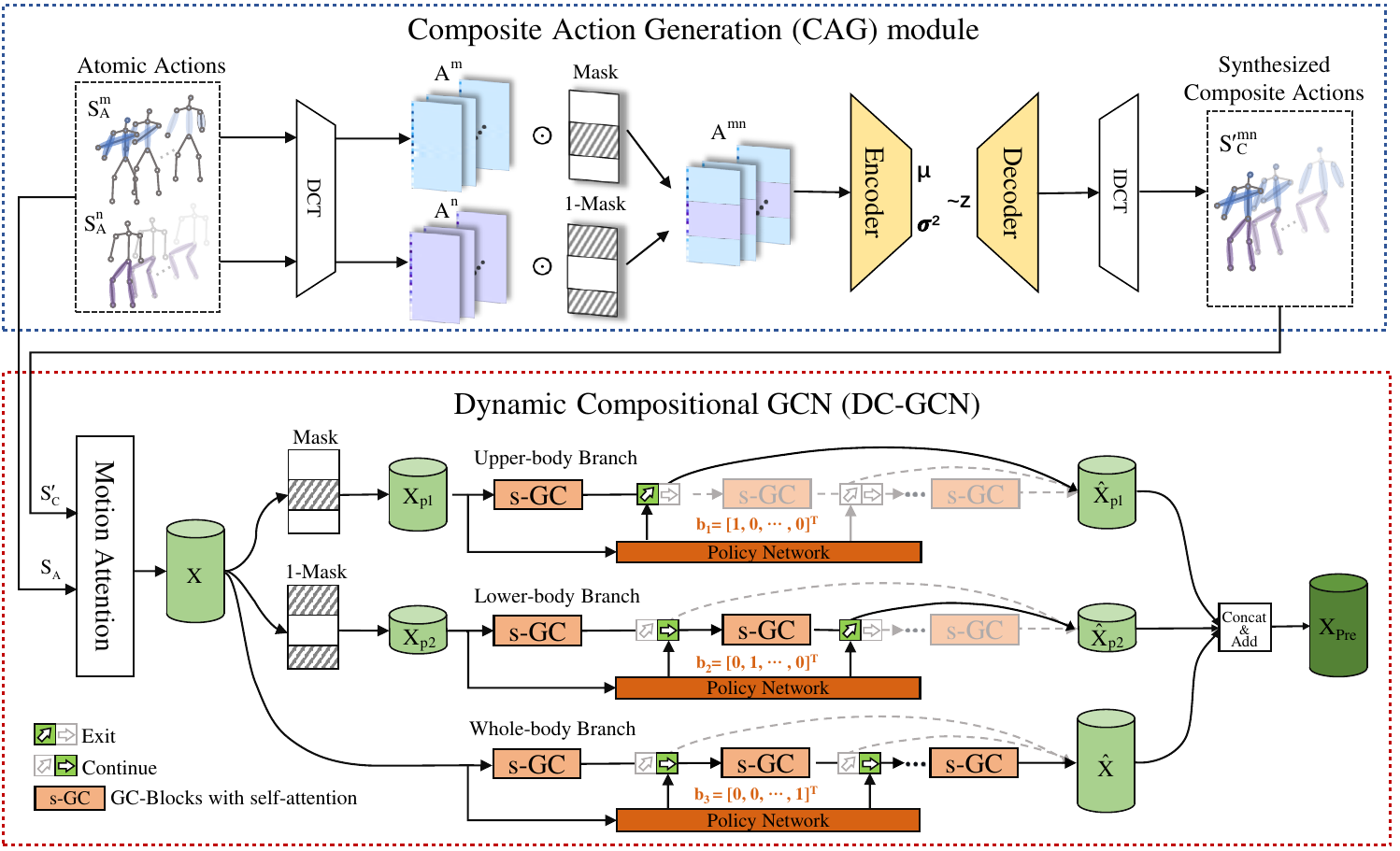}
\vspace{-0.5em}
\caption{Overview of our composite motion prediction framework. \textbf{Composite Action Generation module:} Based on a Variational AutoEncoder, we generate characterized composite actions utilizing only atomic actions. This module is only used during the training procedure. \textbf{Dynamic Compositional GCN:} Each skeleton sequence is first encoded by motion attention to make better use of historical features and then fed to the prediction branches. Two of the three branches model the partial human skeleton while the third branch models the entire human skeleton, while each branch has a policy network that determines where input features terminate the procedure (indicated by the black arrows).}
\vspace{-0.3em}
\label{overview}
\end{figure*}

\section{Our Proposed Framework}
\subsection{Problem Definition}
The task of composite motion prediction is to predict the future motion sequences of a single person. We use $ \mathbf{S}_{1:N} = [\mathbf {s}_1, \mathbf {s}_2, \cdots, \mathbf {s}_N] $ to represent the $N$ frames of each history sequence, and $ \mathbf S_{N+1:N+T} $ as the future $ T $ frames we aim to forecast, where $ \mathbf s_i\in\mathbb{\mathbf R}^{3J}$ is the 3D locations of $J$ human major joints.

We aim to utilize only original atomic action sequences $\mathbf {S}_A$ as training data and use both atomic action sequences $\mathbf {S}_A$ and composite action sequences as testing data. As shown in Figure \ref{overview}, our framework is made up of two modules, the first one is the Composite Action Generation (CAG) module, and the other is the Dynamic Compositional GCN (DC-GCN).
 
The CAG module aims to generate synthesized composite action sequences by feeding atomic action sequences $\mathbf S^{m}_A, \mathbf S^{n}_A \in\mathbb{\mathbf R}^{(N+T) \times (3J)}$ from different atomic actions $m$ and $n$:
 \begin{equation}
   \mathbf S^{\prime}_C = \mathcal{A}(\mathbf S^{m}_A, \mathbf S^{n}_A)
\end{equation}
where $\mathcal{A}$ denotes the CAG modle. $\mathbf S^{\prime}_C \in\mathbb{\mathbf R}^{(N+T) \times (3J)}$ denotes the synthesized composite action sequences.

We then feed the synthesized composite action sequences $\mathbf S^{\prime}_C$ as well as the original atomic action sequences $\mathbf {S}_A$ into DC-GCN denoted by $\mathcal{G}$ for training:
\begin{equation}
   \mathbf S_{N+1:N+T} = \mathcal{G}(\mathbf S_{1:N}),\quad \mathbf S\in\{\mathbf {S}_A,\, \mathbf S^{\prime}_C\}
\end{equation}
Where we use $N$ frames of history sequences as input and forecast the future $T$ frames in both the synthesized composite action sequences $\mathbf S^{\prime}_C$ and the atomic action sequences $\mathbf {S}_A$. In the inference phase, we test all original actions including the composite actions as well as the atomic actions on the trained predictor DC-GCN.

We cover the CAG module $\mathcal{A}$ and the DC-GCN $\mathcal{G}$ in detail in the following subsections.

\subsection{Composite Action Generation}
This module aims to expand the action classes present in the training set and rectify the absence of composite motions, functioning as a data augmentation technique. As discussed, a composite action can be dissected into several atomic actions conducted by non-overlapping body parts that occur concurrently. In the case of our dataset, composite actions are defined by pairing atomic actions. Based on VAE, the CAG module comprises two key steps: model training and motion synthesis.

\textbf{Model Training.} \ In the model training process, we expect that the model can reconstruct the atomic action sequences $\mathbf {S}_A$ as much as possible. We first use the Discrete Cosine Transform (DCT) \citep{akhter2008nonrigid,mao2019learning} operation to extract time characteristics from motion sequences, which benefits the acquisition of continuous motion. Specifically, each atomic action sequence $ \mathbf {S}_A \in\mathbb{\mathbf R}^{(N+T) \times (3J)}$ is encoded as DCT coefficients $ \mathbf A\in\mathbb{\mathbf R}^{F \times (3J)}$ with the formula: $\mathbf A  = \text{DCT}(\mathbf {S}_A)$.

We then use a VAE \citep{kingma2013auto} model to generate $ \mathbf A^{\prime} $ as much like $ \mathbf A $ as possible. Let $q(\mathbf z | \mathbf A)$ denote the distribution of latent variable $\mathbf z$ given $\mathbf A$, $p(\mathbf z)$ represent the probability that $\mathbf z$ is obtained by random sampling from a prior distribution (e.g., Gaussian), and $p(\mathbf A|\mathbf z)$ denote the probability that the decoder outputs $\mathbf A$ when taking $\mathbf z$ as input. 

The encoder $\bm{\mathcal{E}}_{\bm{\phi}}$ and the decoder $\bm{\mathcal{D}}_{\bm{\theta}}$ of VAE model use multi-layer perceptrons (MLPs) to model the two distributions of $q(\mathbf z|\mathbf A)$ and $p(\mathbf A |\mathbf z)$. The parameters of the distributions are estimated by optimizing a loss function that includes both KL divergence and reconstruction loss:
\begin{equation}
\mathcal{L}(\bm\phi,\bm\theta) = -\mathcal{KL}(q_{\bm\phi} (\mathbf z|\mathbf{A})\|p(\mathbf z)) + \int q_{\bm\phi} (\mathbf z|\mathbf A) \ log \ p_{\bm\theta} (\mathbf A|\mathbf z)dz
\end{equation}
During training, the encoder $\bm{\mathcal{E}}_{\bm{\phi}}$ produces the mean $\bm\mu$ and the variance $\bm{\sigma^2} $ in the latent space from the input $ \mathbf A $. The latent representation $\mathbf z$ is then sampled from the normal distribution $\mathcal{N}(\bm\mu,\bm{\sigma^2})$, and the decoder $\bm{\mathcal{D}}_{\bm{\theta}}$ reconstructs $ \mathbf A^{\prime} $ from $\mathbf z$. Lastly, the reconstructed sequence $ \mathbf S^{\prime}_A $ is restored from DCT representation using the Inverse-DCT (IDCT) operation: $\mathbf S^{\prime}_A  = \text{IDCT}(\mathbf A^{\prime})$. In this way, we have the model trained.

\textbf{Motion Synthesis.} \ The next step is motion synthesis, where we aim to acquire new classes of composite actions. A composite action in our CHAMP dataset consists of two atomic actions that occur simultaneously using non-overlapping body parts. Utilizing the VAE model, we fuse atomic actions in pairs to synthesize their corresponding composite actions.
We encode two sequences $ \mathbf S^{m}_A $ and $ \mathbf S^{n}_A $, from different atomic actions $m$ and $n$ via the DCT operation and combine them together with a masking mechanism:
\begin{equation}
\mathbf A^{mn} = \mathbf{M} \odot \text{DCT}(\mathbf S^{m}_A) + (1-\mathbf{M}) \odot \text{DCT}(\mathbf S^{n}_A)
\end{equation}
where $\mathbf{M}$ denotes the mask of the human body skeleton, and $\odot$ denotes the Hadamard product. By employing the masking mechanism, specific locations within the human skeleton for the generation of the two actions can be specified.
The next step involves feeding $\mathbf A^{mn}$ into the encoder $\bm{\mathcal{E}}_{\bm{\phi}}$ to compute the mean $\bm\mu_{mn}$ and variance $\bm{\sigma_{mn}^2} $. The latent representation $\mathbf z_{mn}$ is then sampled from the normal distribution $\mathcal{N}(\bm\mu_{mn},\bm{\sigma_{mn}^2})$. Subsequently, the decoder $\bm{\mathcal{D}}_{\bm{\theta}}$ takes $\mathbf z_{mn}$ as input to generate the DCT representation of composite action. Finally, the synthesized sequence $ \mathbf{S^{\prime}}^{mn}_C $ of the composite action `` $m$ meanwhile $n$'' is restored from its DCT representation using the IDCT operation.

By synthesizing pairs of different atomic actions in this manner, we can generate the sequences of composite actions. This mechanism is employed to synthesize corresponding composite actions and fill gaps that exist in the training set. Specifically, we synthesize 10 upper-body atomic actions and 4 lower-body atomic actions in pairs, resulting in 40 new classes of composite actions.

\subsection{Dynamic Compositional GCN}
We draw inspiration from Hisrep \citep{mao2020history} and leverage graph convolutional network (GCN) to model the spatial dependencies among joints of the human skeleton. Following the motion attention model in Hisrep, we first examine a group of past sub-sequences, measuring the similarity between the newest observed sub-sequence and those from the historical set, to identify the best-suited past sub-sequence. We then represent the sub-sequences as Discrete Cosine Transform (DCT) coefficients and feed them into our DC-GCN. To model spatial dependencies between joint coordinates in different parts of the human skeleton, we develop a three-branch predictor with trainable adjacency matrices of varying sizes. Each branch comprised three s-GC Blocks, with an exit after each block for early termination of training samples. Furthermore, there are two additional layers for encoding and decoding DCT coefficients. 

\textbf{Architecture of s-GC Block.} \
Each s-GC block consists of 8 graph convolutional layers. The graph convolutional layers on different branches adopt different fully connected graphs to model the upper body, lower body, and whole body, respectively. These fully-connected graphs capture critical dependencies between different parts of the human skeleton. Notably, the graph convolutional layers in each branch output matrices in the following form:
\begin{equation}
 \begin{split}
 \mathbf H^{(p+1)} &= \sigma( \mathbf A^{(p)}\mathbf H^{(p)}\mathbf W^{(p)})
 \end{split}
\end{equation}
Where $\mathbf H^{(p+1)}$ represents the convolved output matrix, and $\mathbf H^{(p)}$ can be represented as the input matrices $\mathbf H_u^{(p)}\in\mathbf R^{U \times F}$, $\mathbf H_l^{(p)} \in \mathbf R^{L \times F}$ or $\mathbf H_e^{(p)} \in \mathbf R^{E \times F}$ of three branches, denoting $U$, $L$ or $E$ trajectories with $F$ features. 

The size $U$ and $L$ are determined by the mask of the human body skeleton, while the size of the third branch remains fixed and models the complete human skeleton with $E$ joint coordinates. Each $\mathbf A^{(p)}$ is one of the trainable adjacency matrices in layer $p$, with a unique dimensionality depending on the masks of the branches. These adjacency matrices control the strength of connectivity between nodes in the graph. $\mathbf W^{(p)}\in\mathbf R^{F \times \hat{F}}$ encodes trainable weights with the size of 128 $\times$ 128, and $\sigma(\cdot)$ is the activation function $tanh(\cdot)$. To highlight salient features of every joint coordinate, we introduce a multi-head self-attention module after every four graph convolutional layers in the predictor.

To calculate prediction error, we use the Mean Per Joint Position Error (MPJPE) proposed in \citet{ionescu2013human3}:
\begin{equation}
L_r = \frac{1}{J(N+T)}\sum_{t=1}^{N+T} \sum_{j=1}^{J} \Vert \hat{\mathbf s}_{t,j} - \mathbf s_{t,j} \Vert ^2
\end{equation}
where $\hat{\mathbf s}_{t,j} \in \mathbf R^3$ represent the 3D coordinates of the $j^{th}$ joint of the $t^{th}$ motion pose within the predicted sequence $\hat{\mathbf S}_{1:N+T}$ and $\mathbf s_{t,j} \in \mathbf R^3$ indicates the corresponding ground truth.


\textbf{Early Exit Mechanism.} \
Given that the complexity of movement varies across actions and body parts, we have integrated early exit mechanisms in each branch of the predictor. This approach enables us to efficiently process many simple samples at a shallow layer, while deeper neural networks can be utilized for handling more complicated samples, thereby achieving higher prediction accuracy. We adopt a policy network to determine the exit points for each input predictor, thereby improving the efficiency of the reasoning process by reducing redundant calculations.

In our approach, we set $D$ exits on each branch, and for an input sample $\mathbf X$, the forward propagation of a predictor branch $\mathbf{\mathcal S}$ with $D$ exits is represented by:
\begin{equation}
 \hat{\mathbf X}= \mathbf{\mathcal S}^D \circ \mathbf{\mathcal S}^{D-1} \circ \dots \circ \mathbf{\mathcal S}^1(\mathbf X)
\end{equation}
$\mathbf{\mathcal S}^d$ refers to the graph convolution operation of the $d^{th}$ s-GC block, where $1\le d \le D$, and $\circ$ denotes the composition of operations on different s-GC blocks. The incorporated mechanism of early exit permits the inference to terminate at an intermediate part.

\textbf{Policy networks.} \ The termination exit of each sample is determined by the policy network, which is composed of the policy function $\mathbf {\mathcal P}$ and a Straight-Through (ST) Gumbel Estimator \citep{jang2016categorical}. For each branch of the predictor, the input sample $\mathbf X$ passes through the policy network first, which generates a one-hot vector $ \mathbf{b}_i = [ b_1, \cdots, b_D]^T$ to indicate the selection of one of the $D$ exits in $i^{th}$ branch. The selected exit is represented by the element at the corresponding position of $1$, indicating that the sample exits the network from this exit, whereas the element at any other position is represented by $0$. The process can be represented as follows:
\begin{equation}
 \mathbf b_{i}= \text{Gumbel-Softmax}(\mathbf{\mathcal P_i(\mathbf X)})
\end{equation}
where $\mathbf {\mathcal P_i}$ refers to the policy function in the policy network of the $i^{th}$ branch, which is employed for feature extraction of input $\mathbf X$, and the top-scoring exit can be chosen by the Gumbel-Softmax function. The exit selection of each sample is controlled by an independent policy network in every branch of the predictor.

During training, we observe that the policy network tends to converge to a state where it continuously chooses to exit samples from the shallow output of the network, which results in insufficient training of deeper network layers. 
This imbalance is self-reinforcing because the shallow network has fewer parameters and thus converges faster, leading to the policy network preferring to select the shallow output more often. 
To counter this issue and prevent the network from succumbing to an undesirable local minimum, we incorporate constraints into the loss function inspired by \citep{shazeer2017outrageously}. 

We record the number of samples that chose the same exit location in all branches denoted as $Tendency(\mathbf{E}_d) $, which we acknowledge as the corresponding tendency of the exit location $\mathbf E_d,\ 1\le d \le D$. Additionally, we formulate a specialized loss function $L_{tendency}$:
\begin{equation}
Tendency(\mathbf{E}_d) = \sum\limits_{b_d=1}1
\end{equation}
\begin{equation}
L_{tendency} = w_{tendency} \times  \text{CV}(Tendency(\mathbf{E}_d)) 
\end{equation}
The loss function equals the coefficient of variation of the set of tendency values, multiplied by a hand-tuned scaling factor $w_{tendency}$. These constraints ensure that each output is consistently chosen at a proportional rate before all network layers are entirely trained.

\begin{table*}

\centering
\caption{Prediction results of representative actions on the Human3.6M dataset. Results are shown in 3D joint coordinates at 80ms and 400ms. Comparing our method with other GCN-based approaches, our method exhibits competitive performance on the scales of 80ms and 400ms.}
\vspace{-0.5em}
\scalebox{0.9}{
\begin{tabular}{cc|c|cc|cc|cc|cc} \hline

&  &  & \multicolumn{2}{c|}{walking} &  \multicolumn{2}{c|}{eating} & \multicolumn{2}{c|}{smoking}     & \multicolumn{2}{c}{directions}\\ \hline
\multicolumn{2}{c|}{scenarios} &backbone  & 80ms & 400ms    & 80ms & 400ms   & 80ms & 400ms   & 80ms & 400ms 
 \\ \hline
 siMLPe  \citep{guo2023back} &WACV'23  &MLP
&9.9	&39.6
&5.9	&36.1
&6.5	&36.3
&6.5	&55.8
 \\  \hdashline
Hisrep \citep{mao2020history} &ECCV'20 &GCN
&10.0 	&39.8 
&6.4	&36.2
&7.0	&36.4
&7.4	&56.5
\\
PGBIG  \citep{ma2022progressively} &CVPR'22  &GCN
&11.2	&42.8
&6.5	&36.8
&7.3	&37.5
&7.5	&56.0
\\
\hline 
Ours & &GCN
&\textbf{\textcolor{blue}{9.9}}	&\textbf{\textcolor{blue}{38.0}}
&6.8	&\textbf{\textcolor{blue}{36.1}}
&\textbf{\textcolor{blue}{6.6}}	&\textbf{\textcolor{blue}{34.0}}
&\textbf{\textcolor{blue}{7.2}}	&\textbf{\textcolor{blue}{51.2}}
 \\ \hline
 
 &  & & \multicolumn{2}{c|}{phoning} &  \multicolumn{2}{c|}{sitting} & \multicolumn{2}{c|}{walkingtogether}     & \multicolumn{2}{c}{average}\\ \hline
\multicolumn{2}{c|}{scenarios}  &backbone  & 80ms & 400ms    & 80ms & 400ms   & 80ms & 400ms   & 80ms & 400ms 
 \\ \hline
 siMLPe  \citep{guo2023back} &WACV'23 &MLP
&8.1	&48.6
&8.6	&55.2
&8.4	&41.2
&7.7	&44.7
 \\ \hdashline
Hisrep \citep{mao2020history} &ECCV'20 &GCN
&8.6	&49.2
&9.3	&56.0
&8.9	&41.9
&8.2	&45.1
\\
PGBIG \citep{ma2022progressively} &CVPR'22 &GCN
&8.7	&48.8
&9.1	&54.6
&8.9	&43.8
&8.5	&45.8
\\
 \hline 
Ours &  &GCN
&\textbf{\textcolor{blue}{8.5}}	&\textbf{\textcolor{blue}{48.5}}
&9.2	&\textbf{\textcolor{blue}{54.5}}
&\textbf{\textcolor{blue}{8.7}}	&\textbf{\textcolor{blue}{40.1}}
&\textbf{\textcolor{blue}{8.1}}	&\textbf{\textcolor{blue}{43.2}}
 \\ \hline
\end{tabular}}
\label{result_h36m}
\vspace{-0em}
\end{table*}

\begin{table*}[]
\centering
\caption{Prediction results of representative actions on CHAMP in 3D joint coordinates. Results are shown at 40ms, 100ms, 160ms, and 200ms in the future. Our method outperforms other methods on most actions, particularly those with a greater range (e.g., "squat+wave"), and for longer prediction horizons (200ms).}
\vspace{-0.5em}
\resizebox{\textwidth}{!}{
\begin{tabular}{cc|c|cccc|cccc|cccc|cccc} \hline
&   &  & 40ms & 100ms & 160ms & 200ms  & 40ms & 100ms & 160ms & 200ms   & 40ms & 100ms & 160ms & 200ms   & 40ms & 100ms & 160ms & 200ms      \\ \hline
\multicolumn{2}{c|}{scenarios}  &backbone  & \multicolumn{4}{c|}{squatUp} &  \multicolumn{4}{c|}{standUp} & \multicolumn{4}{c|}{clockwise} & \multicolumn{4}{c}{right}
   \\ \hline
siMLPe \citep{guo2023back} &WACV'23 &MLP
&46.9	&76.5	&95.6	&94.4
&46.9	&65.9	&63.4	&63.3
&33.4	&49.2	&68.5	&81.0
&16.2	&42.7	&89.9	&107.1
 \\ \hdashline
Hisrep \citep{mao2020history} &ECCV'20 &GCN
&44.2	&63.9	&68.4	&68.5
&49.5	&71.2	&76.0	&77.9
&39.9	&60.9	&82.0	&96.8
&19.5	&56.3	&92.0	&102.3
 \\ 
PGBIG \citep{ma2022progressively} &CVPR'22  &GCN
&63.5	&89.0	&82.6	&75.3
&41.0	&66.1	&69.1	&65.7
&30.6	&54.5	&68.6	&77.5
&15.7	&35.6	&51.8	&61.7
 \\ 

Ours &  &GCN
&\textbf{\textcolor{blue}{31.0}}	&\textbf{\textcolor{blue}{43.6}}	&\textbf{\textcolor{blue}{47.0}}	&\textbf{\textcolor{blue}{44.6}}

&41.4	&\textbf{\textcolor{blue}{50.9}}	&\textbf{\textcolor{blue}{49.2}}	&\textbf{\textcolor{blue}{48.5}}

&\textbf{\textcolor{blue}{28.9}}	&\textbf{\textcolor{blue}{44.1}}	&\textbf{\textcolor{blue}{63.8}}	&\textbf{\textcolor{blue}{76.3}}

&\textbf{\textcolor{blue}{14.3}}	&\textbf{\textcolor{blue}{27.7}}	&\textbf{\textcolor{blue}{39.8}}	&\textbf{\textcolor{blue}{47.5}}
 \\ \hline

\multicolumn{2}{c|}{scenarios }  &backbone  & \multicolumn{4}{c|}{standUp+wave} & \multicolumn{4}{c|}{standUp+clockwise} &  \multicolumn{4}{c|}{sitDown+counterclockwise} & \multicolumn{4}{c}{sitDown+keepClose}    \\ \hline

siMLPe \citep{guo2023back} &WACV'23  &MLP
&61.9	&113.7	&140.7	&145.1
&60.5	&95.6	&113.2	&113.8
&62.7	&108.0	&132.8	&142.4
&65.3	&109.0	&131.4	&141.5
\\ \hdashline
Hisrep \citep{mao2020history} &ECCV'20  &GCN
&57.3	&98.2	&122.2	&128.2
&55.6	&81.8	&96.9	&102.3
&59.8	&101.2	&126.9	&136.8
&57.2	&90.5	&114.8	&128.7
 \\ 
PGBIG  \citep{ma2022progressively}  &CVPR'22  &GCN
&57.0	&105.1	&135.2	&147
&59.2	&97.9	&115.6	&123.1
&57.0	&106.5	&142.8	&160.1
&58.2	&108.3	&146.5	&165.1
 \\ 
Ours &  &GCN
&\textbf{\textcolor{blue}{52.6}}	&\textbf{\textcolor{blue}{86.0}}	&\textbf{\textcolor{blue}{102.7}}	&\textbf{\textcolor{blue}{106.1}}

&\textbf{\textcolor{blue}{52.3}}	&\textbf{\textcolor{blue}{72.8}}	&\textbf{\textcolor{blue}{85.2}}	&\textbf{\textcolor{blue}{89.4}}

&\textbf{\textcolor{blue}{56.2}}	&\textbf{\textcolor{blue}{92.3}}	&\textbf{\textcolor{blue}{110.5}}	&\textbf{\textcolor{blue}{117.3}}

&\textbf{\textcolor{blue}{54.0}}	&\textbf{\textcolor{blue}{83.0}}	&\textbf{\textcolor{blue}{101.6}}	&\textbf{\textcolor{blue}{110.3}}
 \\ \hline
 
\multicolumn{2}{c|}{scenarios }  &backbone  & \multicolumn{4}{c|}{sitDown+left} & \multicolumn{4}{c|}{sitDown+wave } &  \multicolumn{4}{c|}{squatUp+keepFar} & \multicolumn{4}{c}{squatUp+nod}    \\ \hline

siMLPe \citep{guo2023back}  &WACV'23  &MLP
&63.6	&112.5	&136.2	&142.2
&60.5	&117.3	&149.4	&159.4
&55.9	&95.8	&113.1	&116.8
&52.6	&76.0	&86.0	&88.2
\\ \hdashline
Hisrep \citep{mao2020history} &ECCV'20  &GCN
&60.3	&103.1	&130.2	&139.9
&54.9	&101.7	&133.2	&145.5
&53.1	&88.9	&106.3	&110.3
&52.5	&84.5	&99.2	&102.7
 \\ 
PGBIG  \citep{ma2022progressively}  &CVPR'22  &GCN
&56.2	&105.4	&140.9	&158.4
&52.9	&104.1	&143.7	&164.2
&62.0	&102.2	&126.8	&132.6
&45.5	&74.3	&94.3	&101.5

 \\ 
Ours &  &GCN
&56.3	&\textbf{\textcolor{blue}{94.6}}	&\textbf{\textcolor{blue}{116.7}}	&\textbf{\textcolor{blue}{123.1}}
&\textbf{\textcolor{blue}{49.1}}	&\textbf{\textcolor{blue}{87.8}}	&\textbf{\textcolor{blue}{110.8}}	&\textbf{\textcolor{blue}{120.8}}
&\textbf{\textcolor{blue}{49.2}}	&\textbf{\textcolor{blue}{77.1}}	&\textbf{\textcolor{blue}{94.8}}	&\textbf{\textcolor{blue}{100.7}}
&50.4	&\textbf{\textcolor{blue}{69.8}}	&\textbf{\textcolor{blue}{75.1}}	&\textbf{\textcolor{blue}{77.5}}
 \\ \hline
 
\multicolumn{2}{c|}{scenarios }  &backbone  & \multicolumn{4}{c|}{squatUp+raiseUp} & \multicolumn{4}{c|}{squatUp+wave} &  \multicolumn{4}{c|}{squat+keepClose} & \multicolumn{4}{c}{squat+left}     \\ \hline
siMLPe \citep{guo2023back}  &WACV'23  &MLP
&51.5	&98.4	&124.1	&122.9
&56.3	&105.3	&137.3	&143.4
&76.1	&126.3	&160.6	&173.2
&73.0	&123.9	&153.3	&164.8
 \\ \hdashline
Hisrep \citep{mao2020history} &ECCV'20  &GCN
&46.9	&85.1	&113.0	&120
&53.5	&91.9	&117.8	&125.5
&70.2	&113.8	&149.4	&160.9
&69.7	&115.8	&145.7	&157.1
 \\ 
PGBIG \citep{ma2022progressively}  &CVPR'22  &GCN
&66.0	&116.5	&132.2	&130.6
&63.3	&102.6	&126.8	&134.0
&63.4	&119.1	&165.1	&186.9
&61.9	&115.6	&156.5	&177.7
 \\ 
Ours &  &GCN
&\textbf{\textcolor{blue}{44.7}}	&\textbf{\textcolor{blue}{74.9}}	&\textbf{\textcolor{blue}{96.1}}	&\textbf{\textcolor{blue}{96.7}}

&\textbf{\textcolor{blue}{49.7}}	&\textbf{\textcolor{blue}{79.3}}	&\textbf{\textcolor{blue}{98.4}}	&\textbf{\textcolor{blue}{107.0}}

&65.9	&\textbf{\textcolor{blue}{103.8}}	&\textbf{\textcolor{blue}{135.5}}	&\textbf{\textcolor{blue}{146.9}}

&64.2	&\textbf{\textcolor{blue}{103.7}}	&\textbf{\textcolor{blue}{131}}	&\textbf{\textcolor{blue}{141.8}}

 \\ \hline

\multicolumn{2}{c|}{scenarios }  &backbone  & \multicolumn{4}{c|}{squat+wave} & \multicolumn{4}{c|}{standUp+counterclockwise} &  \multicolumn{4}{c|}{standUp+raiseUp} & \multicolumn{4}{c}{standUp+shake}    \\ \hline
siMLPe \citep{guo2023back}  &WACV'23   &MLP
&73.9	&139.4	&182.7	&193.9
&60.5	&98.9	&120.8	&122.8
&39.7	&85.6	&115.7	&118.2
&42.7	&63.6	&68.0	&66.7
 \\ \hdashline
Hisrep \citep{mao2020history} &ECCV'20  &GCN
&68.4	&121.9	&158.9	&168.3
&57.7	&85.3	&100.6	&104.7
&39.7	&83.5	&111.8	&115.2
&42.8	&70.2	&79.1	&81.3
 \\ 
PGBIG \citep{ma2022progressively}  &CVPR'22  &GCN
&64.0	&127.7	&175.5	&199.2
&60.2	&98.1	&116.7	&124.7
&53.5	&106.8	&130.9	&135.5
&36.0	&60.1	&73.8	&79.5
 \\ 
Ours &  &GCN
&\textbf{\textcolor{blue}{62.2}}	&\textbf{\textcolor{blue}{106.8}}	&\textbf{\textcolor{blue}{136.7}}	&\textbf{\textcolor{blue}{148.4}}
&\textbf{\textcolor{blue}{54.4}}	&\textbf{\textcolor{blue}{75.1}}	&\textbf{\textcolor{blue}{85.2}}	&\textbf{\textcolor{blue}{87.3}}
&\textbf{\textcolor{blue}{34.7}}	&\textbf{\textcolor{blue}{66.3}}	&\textbf{\textcolor{blue}{89.8}}	&\textbf{\textcolor{blue}{93.9}}
&39.5	&\textbf{\textcolor{blue}{58.4}}	&\textbf{\textcolor{blue}{63.7}}	&\textbf{\textcolor{blue}{61.2}}

 \\ \hline
\end{tabular}}
\label{result_champ}
\end{table*}

\section{Experiments}
Our primary objective is to provide a solution for the composite human motion prediction task. We evaluate the performance of our method on our proposed CHAMP dataset and compare the performance of our method with state-of-the-art approaches in current human motion prediction tasks, including Hisrep \citep{mao2020history}, PGBIG \citep{ma2022progressively} and siMLPe \citep{guo2023back}, which have become baseline methods used in newly published approaches. 
Additionally, we demonstrate the generalizability of our model by presenting results for current human motion prediction tasks, compared with Hisrep, PGBIG, and siMLPe on the Human3.6M dataset \citep{ionescu2013human3}.

\subsection{Datasets and Settings}
\quad \textbf{CHAMP dataset} is a large-scale composite human motion prediction dataset performed by 22 subjects. There are a total of 66 pose classes in the dataset, divided into two main groups, i.e., atomic actions and composite actions. In detail, it contains 16 atomic actions, including 10 upper body actions (raise up, nod, wave, etc.), 5 lower body actions (sit down, squat, walking, etc.), and a still state action. The 50 composite action classes are the pairwise combination of atomic action classes.
For the evaluation of the composite human motion prediction task, we examine 15 classes of atomic actions and 40 classes of composite actions. 15 classes of atomic actions performed by 21 subjects are used as training data, where the labels are specified as: ``still'', ``sitDown'', ``standUp'', ``squat'', ``squatUp'', ``clockwise'', ``counterclockwise'', ``keepClose'', ``keepFar'', ``left'', ``right'', ``nod'', ``shake'', ``wave'', and ``raiseUp''.
Validation data contains 40 classes of composite actions carried out by 11 subjects. Specifically, composite actions performed by subjects 2, 4, 6, 8, 10, 12, 14, 16, 18, 20, and 22 are used as validation data. For the test dataset, we include the remaining composite actions performed by subjects 1, 3, 5, 7, 9, 11, 13, 15, 17, 19, and 21, as well as 15 categories of atomic actions performed by subject 2. Additionally, we follow standard data processing procedures used in other datasets and eliminate the global translation of each human pose.

\begin{figure}
\centering 
\includegraphics[width=0.99\columnwidth]{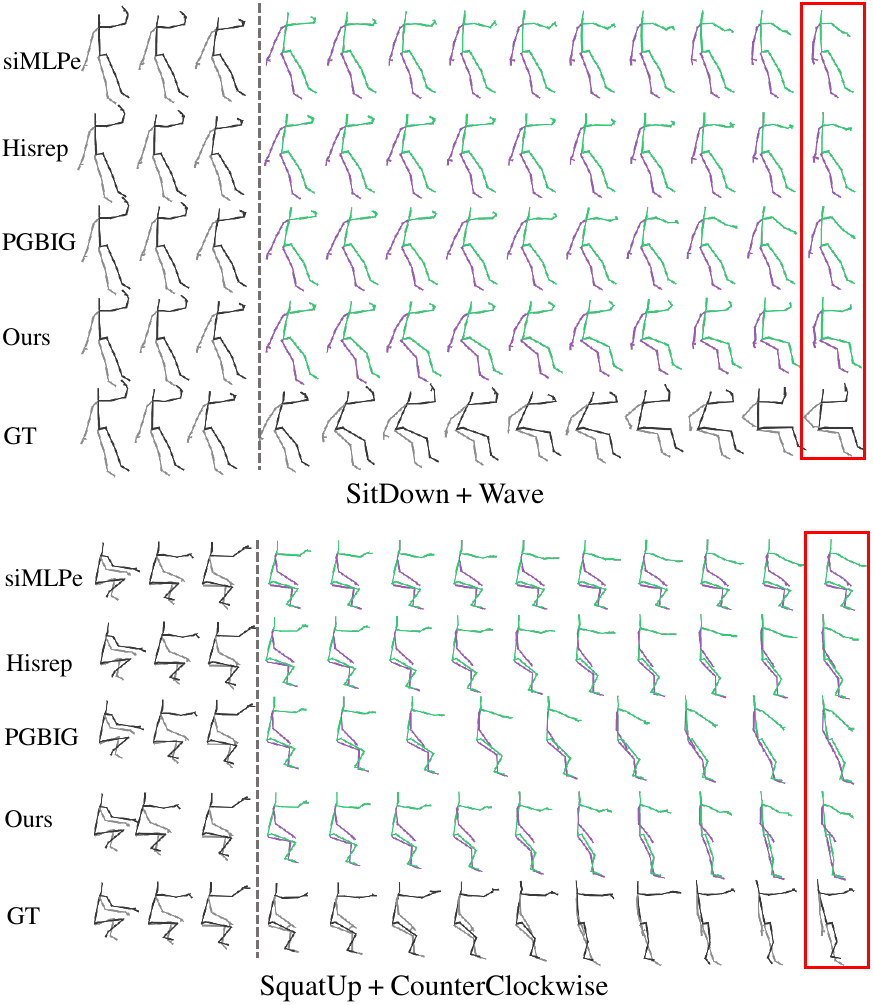}
\vspace{-1em}
\caption{Comparison of our visualization results with Hisrep \citep{mao2020history}, PGBIG \citep{ma2022progressively}, and siMLPe \citep{guo2023back}. Here we use the action ``SitDown + Wave'' and ``squatUp + CounterClockwise'' as examples. The ground truth is shown as black-grey skeletons and the predictions as green-purple. The red boxes mark the frame where our method is closest to ground truth.}
\vspace{-1.1em}
\label{visualization}
\end{figure}

\textbf{Human3.6M dataset \citep{ionescu2013human3}} is the most widely used benchmark dataset for the current human motion prediction task, which uses the same categories of actions in both training and testing. The dataset consists of 7 actors performing 15 actions, with each pose having 32 labeled joints. Following previous work \citep{guo2023back}, we employ subjects S1, S6, S7, S8, and S9 for training, S5 for testing, and S11 for validation. To ensure fair evaluation, we eliminate global translations of each human pose and down-sample the motion sequences to 25 frames per second. For fair comparison, we report our results on 256 samples per action of S5.

\textbf{Implementation details.} We report the results of the 3D joint coordinates on both the CHAMP dataset and the Human3.6M dataset and show the Mean Per Joint Position Error (MPJPE) in millimeters introduced in Section 3.
For the CHAMP dataset, we trained our Composite Action Generation module for 400 epochs and DC-GCN module for 50 epochs. In the DC-GCN module, our approach uses 20 frames of input on the CHAMP dataset and 50 frames on the Human3.6M dataset, while predicting 10 frames of future poses. The learning rate of the CAG module is set to 0.0005, and the learning rate of the DC-GCN is 0.0005 with a 0.96 decay every epoch. The batch size is set to 32 for both the CAG module and the DC-GCN. Our code is in Pytorch and uses ADAM \citep{kingma2014adam} as an optimizer. All the models are trained and tested on an NVIDIA RTX 3080 Ti GPU.

\begin{table}
\centering
\caption{Average prediction results of composite actions and atomic actions in 3D joint coordinates. Results are shown at 40ms, 100ms, 160ms, and 200ms in the future.}
\vspace{-1em}
\resizebox{\columnwidth}{!}{
\begin{tabular}{cc|c|cccc|c} \hline
& &backbone  & 40ms & 100ms & 160ms & 200ms & average       \\ \hline
siMLPe \citep{guo2023back} &WACV'23  &MLP
&50.7	&87.3	&109.8	&115.3 &90.8
\\ \hdashline
Hisrep \citep{mao2020history} &ECCV'20 &GCN
&48.3	&81.7	&104.2	&111.7 &86.5 
\\
PGBIG \citep{ma2022progressively} &CVPR'22  &GCN
&47.7	&86.2	&111.3	&120.9 &91.5
\\

\hline
Ours & &GCN
&\textbf{\textcolor{blue}{44.6}}	&\textbf{\textcolor{blue}{72.8}}	&\textbf{\textcolor{blue}{91.3}}	&\textbf{\textcolor{blue}{96.3 }}  &\textbf{\textcolor{blue}{76.2}}
 \\ \hline
\end{tabular}}
\label{aver_result}
\end{table}

\subsection{Comparison with State-of-the-arts}
We compared our method with Hisrep \citep{mao2020history}, PGBIG \citep{ma2022progressively} and siMLPe \citep{guo2023back} on both the composite human motion prediction task and the current human motion prediction task.

\textbf{Human Motion Prediction Task.} \
We first present the results of our approach in the current human motion prediction task, compared with the state-of-the-art methods on the Human3.6M dataset. As the training and testing for the current human motion prediction task use atomic actions of the same categories and do not require testing corresponding composite actions, there is no need to utilize the Composite Action Generation module for motion synthesis, and we only focus on evaluating the performance of the DC-GCN module on the Human3.6M dataset.

Table \ref{result_h36m} reports the MPJPE results of seven representative actions ``walking'', ``eating'', ``smoking'', ``directions'', ``phoning'', ``sitting'', and ``walkingtogether'' in the short-term forecast. As our approach is a GCN-based method, we compare it primarily with two other GCN-based approaches (Hisrep and PGBIG). As we can see, the performance of our approach on the scale of 400ms is competitive compared with other methods, particularly in the actions of "smoking" and "direction", surpassing the state-of-the-art methods by 2.3mm and 4.6mm. The results show that \uline{our method achieves good performance on current human motion prediction tasks using the Human3.6M dataset.} 
This can be attributed to the multi-branch architecture of the DC-GCN. Many atomic actions in the Human3.6M dataset only involve a large range of movements in either the upper body or lower body, and our DC-GCN models various parts of the body precisely and better extracts key features of atomic actions.

\textbf{Composite Human Motion Prediction Task.}
Table \ref{result_champ} presents the prediction results for representative actions in the CHAMP dataset. Table \ref{aver_result} shows the average results for all 40 classes of composite actions and 15 classes of atomic actions in the CHAMP dataset. Because of the narrow time span of actions in the CHAMP dataset, we predict and show results for the next 200 milliseconds. The ``+'' symbol denotes instances when a person is simultaneously performing two actions.  

It can be seen that Hisrep performs better than PGBIG and siMLPe in the majority of composite actions, and our proposed method outperforms other methods for most actions, except for ``standUp'', ``sitDown + left'', ``squatUp + nod'', ``squat + left'', and ``standUp + shake'' on a scale of 40ms. Moreover, our method works most effectively on actions that require significant amplitude of motion, such as ``standUp+raiseUp'' and ``squat+wave''. These results suggest that our model performs better in actions that involve large movements than in actions that require relatively stationary movements and this advantage becomes more pronounced as the number of forecast frames increases.

Figure \ref{visualization} shows the visualization results of prediction samples ``sitDown + Wave'', and ``squatUp + CounterClockwise''. The visualization of ``sitDown + Wave'' demonstrates how our model outperforms other algorithms in terms of better predictions of both arms and leg movements in the last few frames. In particular, our model displays full bending of legs, consistent with ground truth, while the algorithms show incompletely bent legs. The superior performance of our model could be attributed to two key factors. Firstly, by generating new classes of composite actions from atomic actions, we overcome the issue of inadequate data. 
Secondly, the use of multi-branch models that represent various body parts better extract the critical features of atomic actions and minimize interference from insignificant body parts.

\begin{table}
\centering
\caption{Ablation of different components of our network on the CHAMP dataset. The error is measured in millimeters.}
\vspace{-1em}
\resizebox{\columnwidth}{!}{
\begin{tabular}{ccc|cccc|c} \hline
CAG &EE &C-GCN & 40ms & 100ms & 160ms & 200ms  & average     \\ \hline
\checkmark  & &     &46.3	&78.6	&100.2	&106.1	&82.8 \\
\checkmark & &\checkmark      &44.5	&72.8	&92.7	&98.8	&77.2 \\
 &\checkmark &\checkmark    &45.1	&74.4	&95.0	&101.4	&79.0\\ \hline
\checkmark &\checkmark &\checkmark   &\textbf{\textcolor{blue}{44.6}}	&\textbf{\textcolor{blue}{72.8}}	&\textbf{\textcolor{blue}{91.3}}	&\textbf{\textcolor{blue}{96.3}}	&\textbf{\textcolor{blue}{76.2}}\\ \hline

\end{tabular}}
\label{ablation_component}
\end{table}

\begin{table}
\centering
\caption{Ablations of early exit mechanism on the CHAMP dataset. The error is measured in millimeters.}
\vspace{-1em} 
\resizebox{\columnwidth}{!}{
\begin{tabular}{c|cccc|c} \hline
 & 40ms & 100ms & 160ms & 200ms  & average     \\ \hline
exit 1 &44.9	&74.2	&93.2	&98.7 &77.8\\ 
exit 2	&45.0	&73.5	&91.9	&97.8  &77.1   \\
exit 3    &44.5	&72.8	&92.7	&98.8	&77.2  \\  \hline
constrain 10 &44.7	&73.0	&91.6	&96.9  &76.6\\
constrain 20  &\textbf{\textcolor{blue}{44.6}}	&\textbf{\textcolor{blue}{72.8}}	&\textbf{\textcolor{blue}{91.3}}	&\textbf{\textcolor{blue}{96.3}}  &\textbf{\textcolor{blue}{76.2}} \\
constrain 50 &44.6	&73.6	&92.6	&97.4  &77.1\\ \hline

\end{tabular}}
\label{ablation_ee}
\end{table}

\begin{table}
\centering
\caption{Computational cost with floating-point operations (1e-8) for each early exit mechanism and the basic model on CHAMP and Human3.6M dataset. ``w/o'' denotes the basic model without early exit mechanism, while ``w/'' denotes the model using early exit mechanism.  }
\vspace{-1em}
\resizebox{\columnwidth}{!}{
\begin{tabular}{c|c|c|c|c|c} \hline

\multicolumn{1}{c|}{\multirow{2}*{}} &\multicolumn{3}{c|}{\textbf{CHAMP}} &\multicolumn{2}{c}{\textbf{Human3.6M}}\\
\cline{2-6}
\multicolumn{1}{c|}{}  & w/o   &\tabincell{c}{w/ \\(atomic)}  &\tabincell{c}{w/ \\ (composite)}   & w/o   & w/  \\ \hline

whole-body branch &1.17  &1.10(6.2\%)	&1.17(0.2\%)  &1.05  &0.82(22.0\%)	    \\
upper-body branch &0.71  &0.60(16.2\%)	&0.67(6.3\%)	&0.61  &0.53(11.9\%) \\
lower-body branch &0.32  &0.20(37.4\%)	&0.30(6.3\%)	&0.32  &0.18(42.6\%) \\
 \hline
sum  &2.21	&1.90(14.0\%)  &2.14(3.0\%)  &1.98  &1.54(22.2\%)	\\ \hline

\end{tabular}}
\label{cost}
\vspace{-0.2cm}
\end{table}

\subsection{Ablation Studies} 
\quad \textbf{The importance of Composite Action Generation module.} \ In Table \ref{ablation_component}, we ablate the different components of the network on the CHAMP dataset. One significant difference between our method and other methods is the ability to generate composite actions by the Composite Action Generation module, enabling the prediction model to learn the inherent connections between different body parts of the composite actions in the training process. Removing the Composite Action Generation module has increased errors to 79.0, thereby indicating the effectiveness of the CAG module in generating composite actions. 

\textbf{The importance of early exit mechanism.} \  We also conduct an experiment to evaluate the impact of removing the early exit mechanism from the DC-GCN, resulting in a higher average prediction error from 76.2 to 77.5. It shows that utilizing the early exit mechanism has been shown to enhance prediction accuracy. The results indicate that implementing the early exit mechanism has potential benefits for improving prediction accuracy in predicting composite human actions. This observation can be attributed to the fact that certain body parts tend to remain stationary in the atomic actions, leading to superfluous network computations and loss in prediction accuracy. The early exit mechanism can mitigate such computational redundancies, leading to more accurate predictions.

\textbf{The importance of the DC-GCN architecture.} \ We Replace the DC-GCN module with the universal GCN that is commonly applied \citep{mao2019learning,mao2020history}. The results show that switching from the DC-GCN to the universal GCN yields a sharp increase in the average prediction error from 76.2 to 82.8 on the CHAMP dataset, which verifies that the Dynamic Compositional GCN plays a critical role in enhancing the model's prediction accuracy.

\textbf{Ablation on early exit mechanism.} \ Table \ref{ablation_ee} shows the impact of exit point selection and applying constraints on prediction results. We aim to determine whether incorporating the early exit mechanism would improve prediction accuracy. We conduct an experiment in which all samples are only allowed to exit at the first, second, or third exit point in the branches of the DC-GCN. The experimental results demonstrate that the prediction accuracy of dynamic network structures with early exit mechanisms exceeds that of fixed network structures. To fully train all network layers, we introduce $L_{tendency}$ to the loss function to enforce constraints, which imposes constraints and enables the policy networks to evenly select the constrained exits. We impose constraints on various training epochs, ultimately opting to constrain the policy network's selection during the initial 20 training epochs, after which it could autonomously learn to choose exits for each sample throughout the rest of the training process.

\textbf{Computational Efficiency.} \ 
In the three sub-networks of DC-GCN, we employ an early exit mechanism. By significantly reducing computational costs, this method effectively maintains a balance between model accuracy and computational efficiency. The extent of saved computing depends on the features of the dataset and the model's complexity. Consequently, each dataset is evaluated to estimate the computational savings achieved by the early exit mechanism on each branch. Table \ref{cost} shows the computational cost with floating-point operations (FLOPs), it illustrates that atomic actions conserve more computational resources than composite actions on the CHAMP dataset, validating our hypothesis that atomic actions are simpler to predict than composite ones. The Human3.6M dataset records a higher percentage of resource conservation than the CHAMP dataset, conceivably, primarily due to the intricacy of CHAMP's movements in comparison to Human3.6M. The savings in computation vary across branches, which we attribute to the varying complexity of movements in different body parts. Specifically, movements in the lower body tend to be less complex than those in the upper body.
\vspace{-0.15cm}


\section{Conclusion}
\noindent This paper addresses the challenge of predicting composite human motion, which is more complicated than mainstream atomic action-based human motion prediction. To support future research in this field, we collect the CHAMP dataset, which comprises 16 atomic actions and 50 composite actions. Our approach to this task involves using an efficient framework comprising a Composite Action
Generation (CAG) module and Dynamic Compositional GCN (DC-GCN). The CAG module addresses the scarcity of composite training data by using atomic actions to synthesize composite actions. And the DC-GCN models both the partial and entire human skeleton and forecasts future poses using the GCN-based network. Moreover, We incorporate an early exit mechanism into our prediction framework, which effectively balances prediction accuracy and computational efficiency. Our method surpasses the performance of the baseline approaches (PGBIG \citep{ma2022progressively}, Hisrep \citep{mao2020history}, and siMLPe \citep{guo2023back}) in a significant manner. 

\begin{acks}
This work was supported by the National Natural Science Foundation of China (No. 62203476).

\end{acks}

\vfill
\newpage
\normalem
\bibliographystyle{ACM-Reference-Format}
\balance
\bibliography{main}


\begin{thebibliography}{36}


\ifx \showCODEN    \undefined \def \showCODEN     #1{\unskip}     \fi
\ifx \showDOI      \undefined \def \showDOI       #1{#1}\fi
\ifx \showISBNx    \undefined \def \showISBNx     #1{\unskip}     \fi
\ifx \showISBNxiii \undefined \def \showISBNxiii  #1{\unskip}     \fi
\ifx \showISSN     \undefined \def \showISSN      #1{\unskip}     \fi
\ifx \showLCCN     \undefined \def \showLCCN      #1{\unskip}     \fi
\ifx \shownote     \undefined \def \shownote      #1{#1}          \fi
\ifx \showarticletitle \undefined \def \showarticletitle #1{#1}   \fi
\ifx \showURL      \undefined \def \showURL       {\relax}        \fi
\providecommand\bibfield[2]{#2}
\providecommand\bibinfo[2]{#2}
\providecommand\natexlab[1]{#1}
\providecommand\showeprint[2][]{arXiv:#2}

\bibitem[Akhter et~al\mbox{.}(2008)]%
        {akhter2008nonrigid}
\bibfield{author}{\bibinfo{person}{Ijaz Akhter}, \bibinfo{person}{Yaser
  Sheikh}, \bibinfo{person}{Sohaib Khan}, {and} \bibinfo{person}{Takeo
  Kanade}.} \bibinfo{year}{2008}\natexlab{}.
\newblock \showarticletitle{Nonrigid structure from motion in trajectory
  space}.
\newblock \bibinfo{journal}{\emph{Advances in neural information processing
  systems}}  \bibinfo{volume}{21} (\bibinfo{year}{2008}).
\newblock


\bibitem[Aksan et~al\mbox{.}(2021)]%
        {aksan2021spatio}
\bibfield{author}{\bibinfo{person}{Emre Aksan}, \bibinfo{person}{Manuel
  Kaufmann}, \bibinfo{person}{Peng Cao}, {and} \bibinfo{person}{Otmar
  Hilliges}.} \bibinfo{year}{2021}\natexlab{}.
\newblock \showarticletitle{A spatio-temporal transformer for 3d human motion
  prediction}. In \bibinfo{booktitle}{\emph{International Conference on 3D
  Vision (3DV)}}. \bibinfo{pages}{565--574}.
\newblock


\bibitem[Bolukbasi et~al\mbox{.}(2017)]%
        {bolukbasi2017adaptive}
\bibfield{author}{\bibinfo{person}{Tolga Bolukbasi}, \bibinfo{person}{Joseph
  Wang}, \bibinfo{person}{Ofer Dekel}, {and} \bibinfo{person}{Venkatesh
  Saligrama}.} \bibinfo{year}{2017}\natexlab{}.
\newblock \showarticletitle{Adaptive neural networks for efficient inference}.
  In \bibinfo{booktitle}{\emph{International Conference on Machine Learning}}.
  PMLR, \bibinfo{pages}{527--536}.
\newblock


\bibitem[Corona et~al\mbox{.}(2020)]%
        {corona2020context}
\bibfield{author}{\bibinfo{person}{Enric Corona}, \bibinfo{person}{Albert
  Pumarola}, \bibinfo{person}{Guillem Alenya}, {and} \bibinfo{person}{Francesc
  Moreno-Noguer}.} \bibinfo{year}{2020}\natexlab{}.
\newblock \showarticletitle{Context-aware human motion prediction}. In
  \bibinfo{booktitle}{\emph{IEEE/CVF Conference on Computer Vision and Pattern
  Recognition (CVPR)}}. \bibinfo{pages}{6992--7001}.
\newblock


\bibitem[Cui and Sun(2021)]%
        {cui2021towards}
\bibfield{author}{\bibinfo{person}{Qiongjie Cui} {and}
  \bibinfo{person}{Huaijiang Sun}.} \bibinfo{year}{2021}\natexlab{}.
\newblock \showarticletitle{Towards accurate 3d human motion prediction from
  incomplete observations}. In \bibinfo{booktitle}{\emph{Proceedings of the
  IEEE/CVF Conference on Computer Vision and Pattern Recognition}}.
  \bibinfo{pages}{4801--4810}.
\newblock


\bibitem[Dang et~al\mbox{.}(2021)]%
        {dang2021msr}
\bibfield{author}{\bibinfo{person}{Lingwei Dang}, \bibinfo{person}{Yongwei
  Nie}, \bibinfo{person}{Chengjiang Long}, \bibinfo{person}{Qing Zhang}, {and}
  \bibinfo{person}{Guiqing Li}.} \bibinfo{year}{2021}\natexlab{}.
\newblock \showarticletitle{MSR-GCN: Multi-scale residual graph convolution
  networks for human motion prediction}. In \bibinfo{booktitle}{\emph{IEEE/CVF
  International Conference on Computer Vision (CVPR)}}.
  \bibinfo{pages}{11467--11476}.
\newblock


\bibitem[Fragkiadaki et~al\mbox{.}(2015)]%
        {fragkiadaki2015recurrent}
\bibfield{author}{\bibinfo{person}{Katerina Fragkiadaki},
  \bibinfo{person}{Sergey Levine}, \bibinfo{person}{Panna Felsen}, {and}
  \bibinfo{person}{Jitendra Malik}.} \bibinfo{year}{2015}\natexlab{}.
\newblock \showarticletitle{Recurrent network models for human dynamics}. In
  \bibinfo{booktitle}{\emph{IEEE International Conference on Computer Vision
  (ICCV)}}. \bibinfo{pages}{4346--4354}.
\newblock


\bibitem[Graves(2016)]%
        {graves2016adaptive}
\bibfield{author}{\bibinfo{person}{Alex Graves}.}
  \bibinfo{year}{2016}\natexlab{}.
\newblock \showarticletitle{Adaptive computation time for recurrent neural
  networks}.
\newblock \bibinfo{journal}{\emph{arXiv preprint arXiv:1603.08983}}
  (\bibinfo{year}{2016}).
\newblock


\bibitem[Guo et~al\mbox{.}(2023)]%
        {guo2023back}
\bibfield{author}{\bibinfo{person}{Wen Guo}, \bibinfo{person}{Yuming Du},
  \bibinfo{person}{Xi Shen}, \bibinfo{person}{Vincent Lepetit},
  \bibinfo{person}{Xavier Alameda-Pineda}, {and} \bibinfo{person}{Francesc
  Moreno-Noguer}.} \bibinfo{year}{2023}\natexlab{}.
\newblock \showarticletitle{Back to mlp: A simple baseline for human motion
  prediction}. In \bibinfo{booktitle}{\emph{Proceedings of the IEEE/CVF Winter
  Conference on Applications of Computer Vision}}. \bibinfo{pages}{4809--4819}.
\newblock


\bibitem[Huang et~al\mbox{.}(2017)]%
        {huang2017multi}
\bibfield{author}{\bibinfo{person}{Gao Huang}, \bibinfo{person}{Danlu Chen},
  \bibinfo{person}{Tianhong Li}, \bibinfo{person}{Felix Wu},
  \bibinfo{person}{Laurens Van Der~Maaten}, {and} \bibinfo{person}{Kilian~Q
  Weinberger}.} \bibinfo{year}{2017}\natexlab{}.
\newblock \showarticletitle{Multi-scale dense networks for resource efficient
  image classification}.
\newblock \bibinfo{journal}{\emph{arXiv preprint arXiv:1703.09844}}
  (\bibinfo{year}{2017}).
\newblock


\bibitem[Ionescu et~al\mbox{.}(2013)]%
        {ionescu2013human3}
\bibfield{author}{\bibinfo{person}{Catalin Ionescu}, \bibinfo{person}{Dragos
  Papava}, \bibinfo{person}{Vlad Olaru}, {and} \bibinfo{person}{Cristian
  Sminchisescu}.} \bibinfo{year}{2013}\natexlab{}.
\newblock \showarticletitle{Human3. 6m: Large scale datasets and predictive
  methods for 3d human sensing in natural environments}.
\newblock \bibinfo{journal}{\emph{IEEE Transactions on Pattern Analysis and
  Machine Intelligence}} \bibinfo{volume}{36}, \bibinfo{number}{7}
  (\bibinfo{year}{2013}), \bibinfo{pages}{1325--1339}.
\newblock


\bibitem[Iwana and Uchida(2021)]%
        {iwana2021empirical}
\bibfield{author}{\bibinfo{person}{Brian~Kenji Iwana} {and}
  \bibinfo{person}{Seiichi Uchida}.} \bibinfo{year}{2021}\natexlab{}.
\newblock \showarticletitle{An empirical survey of data augmentation for time
  series classification with neural networks}.
\newblock \bibinfo{journal}{\emph{Plos one}} \bibinfo{volume}{16},
  \bibinfo{number}{7} (\bibinfo{year}{2021}), \bibinfo{pages}{e0254841}.
\newblock


\bibitem[Jang et~al\mbox{.}(2016)]%
        {jang2016categorical}
\bibfield{author}{\bibinfo{person}{Eric Jang}, \bibinfo{person}{Shixiang Gu},
  {and} \bibinfo{person}{Ben Poole}.} \bibinfo{year}{2016}\natexlab{}.
\newblock \showarticletitle{Categorical reparameterization with
  gumbel-softmax}. In \bibinfo{booktitle}{\emph{arXiv:1611.01144}}.
\newblock


\bibitem[Jie et~al\mbox{.}(2019)]%
        {jie2019anytime}
\bibfield{author}{\bibinfo{person}{Zequn Jie}, \bibinfo{person}{Peng Sun},
  \bibinfo{person}{Xin Li}, \bibinfo{person}{Jiashi Feng}, {and}
  \bibinfo{person}{Wei Liu}.} \bibinfo{year}{2019}\natexlab{}.
\newblock \showarticletitle{Anytime recognition with routing convolutional
  networks}.
\newblock \bibinfo{journal}{\emph{IEEE transactions on pattern analysis and
  machine intelligence}} \bibinfo{volume}{43}, \bibinfo{number}{6}
  (\bibinfo{year}{2019}), \bibinfo{pages}{1875--1886}.
\newblock


\bibitem[Kingma and Ba(2014)]%
        {kingma2014adam}
\bibfield{author}{\bibinfo{person}{Diederik~P Kingma} {and}
  \bibinfo{person}{Jimmy Ba}.} \bibinfo{year}{2014}\natexlab{}.
\newblock \showarticletitle{Adam: A method for stochastic optimization}. In
  \bibinfo{booktitle}{\emph{arXiv:1412.6980}}.
\newblock


\bibitem[Kingma and Welling(2013)]%
        {kingma2013auto}
\bibfield{author}{\bibinfo{person}{Diederik~P Kingma} {and}
  \bibinfo{person}{Max Welling}.} \bibinfo{year}{2013}\natexlab{}.
\newblock \showarticletitle{Auto-encoding variational bayes}.
\newblock \bibinfo{journal}{\emph{arXiv preprint arXiv:1312.6114}}
  (\bibinfo{year}{2013}).
\newblock


\bibitem[Krizhevsky et~al\mbox{.}(2017)]%
        {krizhevsky2017imagenet}
\bibfield{author}{\bibinfo{person}{Alex Krizhevsky}, \bibinfo{person}{Ilya
  Sutskever}, {and} \bibinfo{person}{Geoffrey~E Hinton}.}
  \bibinfo{year}{2017}\natexlab{}.
\newblock \showarticletitle{Imagenet classification with deep convolutional
  neural networks}.
\newblock \bibinfo{journal}{\emph{Commun. ACM}} \bibinfo{volume}{60},
  \bibinfo{number}{6} (\bibinfo{year}{2017}), \bibinfo{pages}{84--90}.
\newblock


\bibitem[Li et~al\mbox{.}(2020b)]%
        {li2020predicting}
\bibfield{author}{\bibinfo{person}{Anliang Li}, \bibinfo{person}{Shuang Wang},
  \bibinfo{person}{Wenzhu Li}, \bibinfo{person}{Shengnan Liu}, {and}
  \bibinfo{person}{Siyuan Zhang}.} \bibinfo{year}{2020}\natexlab{b}.
\newblock \showarticletitle{Predicting human mobility with federated learning}.
  In \bibinfo{booktitle}{\emph{28th International Conference on Advances in
  Geographic Information Systems}}. \bibinfo{pages}{441--444}.
\newblock


\bibitem[Li et~al\mbox{.}(2020a)]%
        {li2020dynamic}
\bibfield{author}{\bibinfo{person}{Maosen Li}, \bibinfo{person}{Siheng Chen},
  \bibinfo{person}{Yangheng Zhao}, \bibinfo{person}{Ya Zhang},
  \bibinfo{person}{Yanfeng Wang}, {and} \bibinfo{person}{Qi Tian}.}
  \bibinfo{year}{2020}\natexlab{a}.
\newblock \showarticletitle{Dynamic multiscale graph neural networks for 3d
  skeleton based human motion prediction}. In
  \bibinfo{booktitle}{\emph{Proceedings of the IEEE/CVF Conference on Computer
  Vision and Pattern Recognition}}. \bibinfo{pages}{214--223}.
\newblock


\bibitem[Liu et~al\mbox{.}(2017)]%
        {liu2017enhanced}
\bibfield{author}{\bibinfo{person}{Mengyuan Liu}, \bibinfo{person}{Hong Liu},
  {and} \bibinfo{person}{Chen Chen}.} \bibinfo{year}{2017}\natexlab{}.
\newblock \showarticletitle{Enhanced skeleton visualization for view invariant
  human action recognition}.
\newblock \bibinfo{journal}{\emph{Pattern Recognition}}  \bibinfo{volume}{68}
  (\bibinfo{year}{2017}), \bibinfo{pages}{346--362}.
\newblock


\bibitem[Liu and Yuan(2018)]%
        {liu2018recognizing}
\bibfield{author}{\bibinfo{person}{Mengyuan Liu} {and} \bibinfo{person}{Junsong
  Yuan}.} \bibinfo{year}{2018}\natexlab{}.
\newblock \showarticletitle{Recognizing human actions as the evolution of pose
  estimation maps}. In \bibinfo{booktitle}{\emph{IEEE/CVF Conference on
  Computer Vision and Pattern Recognition workshops (CVPR)}}.
  \bibinfo{pages}{1159--1168}.
\newblock


\bibitem[Lopes et~al\mbox{.}(2019)]%
        {lopes2019improving}
\bibfield{author}{\bibinfo{person}{Raphael~Gontijo Lopes},
  \bibinfo{person}{Dong Yin}, \bibinfo{person}{Ben Poole},
  \bibinfo{person}{Justin Gilmer}, {and} \bibinfo{person}{Ekin~D Cubuk}.}
  \bibinfo{year}{2019}\natexlab{}.
\newblock \showarticletitle{Improving robustness without sacrificing accuracy
  with patch gaussian augmentation}. In
  \bibinfo{booktitle}{\emph{arXiv:1906.02611}}.
\newblock


\bibitem[Ma et~al\mbox{.}(2022)]%
        {ma2022progressively}
\bibfield{author}{\bibinfo{person}{Tiezheng Ma}, \bibinfo{person}{Yongwei Nie},
  \bibinfo{person}{Chengjiang Long}, \bibinfo{person}{Qing Zhang}, {and}
  \bibinfo{person}{Guiqing Li}.} \bibinfo{year}{2022}\natexlab{}.
\newblock \showarticletitle{Progressively Generating Better Initial Guesses
  Towards Next Stages for High-Quality Human Motion Prediction}. In
  \bibinfo{booktitle}{\emph{Proceedings of the IEEE/CVF Conference on Computer
  Vision and Pattern Recognition}}. \bibinfo{pages}{6437--6446}.
\newblock


\bibitem[Maeda and Ukita(2022)]%
        {maeda2022motionaug}
\bibfield{author}{\bibinfo{person}{Takahiro Maeda} {and}
  \bibinfo{person}{Norimichi Ukita}.} \bibinfo{year}{2022}\natexlab{}.
\newblock \showarticletitle{MotionAug: Augmentation with Physical Correction
  for Human Motion Prediction}. In \bibinfo{booktitle}{\emph{Proceedings of the
  IEEE/CVF Conference on Computer Vision and Pattern Recognition}}.
  \bibinfo{pages}{6427--6436}.
\newblock


\bibitem[Mao et~al\mbox{.}(2020)]%
        {mao2020history}
\bibfield{author}{\bibinfo{person}{Wei Mao}, \bibinfo{person}{Miaomiao Liu},
  {and} \bibinfo{person}{Mathieu Salzmann}.} \bibinfo{year}{2020}\natexlab{}.
\newblock \showarticletitle{History repeats itself: Human motion prediction via
  motion attention}. In \bibinfo{booktitle}{\emph{European Conference on
  Computer Vision (ECCV)}}. \bibinfo{pages}{474--489}.
\newblock


\bibitem[Mao et~al\mbox{.}(2019)]%
        {mao2019learning}
\bibfield{author}{\bibinfo{person}{Wei Mao}, \bibinfo{person}{Miaomiao Liu},
  \bibinfo{person}{Mathieu Salzmann}, {and} \bibinfo{person}{Hongdong Li}.}
  \bibinfo{year}{2019}\natexlab{}.
\newblock \showarticletitle{Learning trajectory dependencies for human motion
  prediction}. In \bibinfo{booktitle}{\emph{IEEE/CVF International Conference
  on Computer Vision (ICCV)}}. \bibinfo{pages}{9489--9497}.
\newblock


\bibitem[Mao et~al\mbox{.}(2021)]%
        {mao2021multi}
\bibfield{author}{\bibinfo{person}{Wei Mao}, \bibinfo{person}{Miaomiao Liu},
  \bibinfo{person}{Mathieu Salzmann}, {and} \bibinfo{person}{Hongdong Li}.}
  \bibinfo{year}{2021}\natexlab{}.
\newblock \showarticletitle{Multi-level motion attention for human motion
  prediction}.
\newblock \bibinfo{journal}{\emph{International Journal of Computer Vision}}
  \bibinfo{volume}{129}, \bibinfo{number}{9} (\bibinfo{year}{2021}),
  \bibinfo{pages}{2513--2535}.
\newblock


\bibitem[McGill and Perona(2017)]%
        {mcgill2017deciding}
\bibfield{author}{\bibinfo{person}{Mason McGill} {and} \bibinfo{person}{Pietro
  Perona}.} \bibinfo{year}{2017}\natexlab{}.
\newblock \showarticletitle{Deciding how to decide: Dynamic routing in
  artificial neural networks}. In \bibinfo{booktitle}{\emph{International
  Conference on Machine Learning}}. PMLR, \bibinfo{pages}{2363--2372}.
\newblock


\bibitem[Shazeer et~al\mbox{.}(2017)]%
        {shazeer2017outrageously}
\bibfield{author}{\bibinfo{person}{Noam Shazeer}, \bibinfo{person}{Azalia
  Mirhoseini}, \bibinfo{person}{Krzysztof Maziarz}, \bibinfo{person}{Andy
  Davis}, \bibinfo{person}{Quoc Le}, \bibinfo{person}{Geoffrey Hinton}, {and}
  \bibinfo{person}{Jeff Dean}.} \bibinfo{year}{2017}\natexlab{}.
\newblock \showarticletitle{Outrageously large neural networks: The
  sparsely-gated mixture-of-experts layer}.
\newblock \bibinfo{journal}{\emph{arXiv preprint arXiv:1701.06538}}
  (\bibinfo{year}{2017}).
\newblock


\bibitem[Tanner and Wong(1987)]%
        {tanner1987calculation}
\bibfield{author}{\bibinfo{person}{Martin~A Tanner} {and}
  \bibinfo{person}{Wing~Hung Wong}.} \bibinfo{year}{1987}\natexlab{}.
\newblock \showarticletitle{The calculation of posterior distributions by data
  augmentation}.
\newblock \bibinfo{journal}{\emph{Journal of the American statistical
  Association}} \bibinfo{volume}{82}, \bibinfo{number}{398}
  (\bibinfo{year}{1987}), \bibinfo{pages}{528--540}.
\newblock


\bibitem[Teerapittayanon et~al\mbox{.}(2016)]%
        {teerapittayanon2016branchynet}
\bibfield{author}{\bibinfo{person}{Surat Teerapittayanon},
  \bibinfo{person}{Bradley McDanel}, {and} \bibinfo{person}{Hsiang-Tsung
  Kung}.} \bibinfo{year}{2016}\natexlab{}.
\newblock \showarticletitle{Branchynet: Fast inference via early exiting from
  deep neural networks}. In \bibinfo{booktitle}{\emph{2016 23rd International
  Conference on Pattern Recognition (ICPR)}}. IEEE,
  \bibinfo{pages}{2464--2469}.
\newblock


\bibitem[Vaswani et~al\mbox{.}(2017)]%
        {vaswani2017attention}
\bibfield{author}{\bibinfo{person}{Ashish Vaswani}, \bibinfo{person}{Noam
  Shazeer}, \bibinfo{person}{Niki Parmar}, \bibinfo{person}{Jakob Uszkoreit},
  \bibinfo{person}{Llion Jones}, \bibinfo{person}{Aidan~N Gomez},
  \bibinfo{person}{{\L}ukasz Kaiser}, {and} \bibinfo{person}{Illia
  Polosukhin}.} \bibinfo{year}{2017}\natexlab{}.
\newblock \showarticletitle{Attention is all you need}.
\newblock \bibinfo{journal}{\emph{Advances in Neural Information Processing
  Systems}}  \bibinfo{volume}{30} (\bibinfo{year}{2017}).
\newblock


\bibitem[Wang et~al\mbox{.}(2021)]%
        {wang2021spatio}
\bibfield{author}{\bibinfo{person}{Ning Wang}, \bibinfo{person}{Guangming Zhu},
  \bibinfo{person}{Liang Zhang}, \bibinfo{person}{Peiyi Shen},
  \bibinfo{person}{Hongsheng Li}, {and} \bibinfo{person}{Cong Hua}.}
  \bibinfo{year}{2021}\natexlab{}.
\newblock \showarticletitle{Spatio-Temporal Interaction Graph Parsing Networks
  for Human-Object Interaction Recognition}. In
  \bibinfo{booktitle}{\emph{Proceedings of the 29th ACM International
  Conference on Multimedia}}. \bibinfo{pages}{4985--4993}.
\newblock


\bibitem[Zhu et~al\mbox{.}(2021)]%
        {zhu2021recurrent}
\bibfield{author}{\bibinfo{person}{Guangming Zhu}, \bibinfo{person}{Lu Yang},
  \bibinfo{person}{Liang Zhang}, \bibinfo{person}{Peiyi Shen}, {and}
  \bibinfo{person}{Juan Song}.} \bibinfo{year}{2021}\natexlab{}.
\newblock \showarticletitle{Recurrent Graph Convolutional Networks for
  Skeleton-based Action Recognition}. In \bibinfo{booktitle}{\emph{2020 25th
  International Conference on Pattern Recognition (ICPR)}}. IEEE,
  \bibinfo{pages}{1352--1359}.
\newblock


\bibitem[Zhu et~al\mbox{.}(2020)]%
        {zhu2020topology}
\bibfield{author}{\bibinfo{person}{Guangming Zhu}, \bibinfo{person}{Liang
  Zhang}, \bibinfo{person}{Hongsheng Li}, \bibinfo{person}{Peiyi Shen},
  \bibinfo{person}{Syed Afaq~Ali Shah}, {and} \bibinfo{person}{Mohammed
  Bennamoun}.} \bibinfo{year}{2020}\natexlab{}.
\newblock \showarticletitle{Topology-learnable graph convolution for
  skeleton-based action recognition}.
\newblock \bibinfo{journal}{\emph{Pattern Recognition Letters}}
  \bibinfo{volume}{135} (\bibinfo{year}{2020}), \bibinfo{pages}{286--292}.
\newblock


\bibitem[Zoph et~al\mbox{.}(2018)]%
        {zoph2018learning}
\bibfield{author}{\bibinfo{person}{Barret Zoph}, \bibinfo{person}{Vijay
  Vasudevan}, \bibinfo{person}{Jonathon Shlens}, {and} \bibinfo{person}{Quoc~V
  Le}.} \bibinfo{year}{2018}\natexlab{}.
\newblock \showarticletitle{Learning transferable architectures for scalable
  image recognition}. In \bibinfo{booktitle}{\emph{Proceedings of the IEEE
  conference on computer vision and pattern recognition}}.
  \bibinfo{pages}{8697--8710}.
\newblock


\end{thebibliography}
\end{document}